\documentclass[10pt,conference,a4paper]{IEEEtran}

\usepackage{graphicx}
\usepackage[caption=false,font=footnotesize,labelfont=sf,textfont=sf]{subfig}
\graphicspath{{figure/}}
\DeclareGraphicsExtensions{.eps, .pdf}
\usepackage{color}		
\usepackage{amsmath,amssymb,amsfonts}
\usepackage{upgreek}
\usepackage{cite}
\usepackage{url}
\usepackage{bm}
\usepackage{booktabs}
\usepackage{multirow}
\usepackage{diagbox}
\usepackage{algorithm}
\usepackage{algorithmic}

\usepackage[bookmarks,colorlinks,unicode]{hyperref}
\hypersetup{
	pdftitle={Low-Rank Tensor Completion by Truncated Nuclear Norm Regularization},
	pdfauthor={Shengke Xue, Wenyuan Qiu, Fan Liu, and Xinyu Jin},
	pdfsubject={2018 24th International Conference on Pattern Recognition (ICPR), August 20-24, 2018, Beijing, China},
	pdfkeywords={Inpainting; Image processing and analysis; Video processing and analysis},
}

\newcommand{\trace}[1]{\text{tr}({#1})}
\newcommand{\norm}[1]{||{#1}||}
\newcommand{\transpose}{{\text{T}}}
\newcommand{\fro}{\text{F}}
\newcommand{\tensor}[1]{\bm{\mathcal{#1}}}
\newcommand{\ftensor}[1]{\bm{\mathcal{\bar{#1}}}}
\newcommand{\fmat}[1]{\bm{\bar{#1}}}
\newcommand{\Rdim}[1]{\in\mathbb{R}^{#1}}
\newcommand{\Cdim}[1]{\in\mathbb{C}^{#1}}
\newcommand{\etal}{\emph{et al}.}

\newcommand{\letter}{\includegraphics[scale=1]{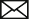}}

\newtheorem{definition}{Definition}[section]
\newtheorem{theorem}[definition]{Theorem}

\begin{document}
%
\title{Low-Rank Tensor Completion by\\ Truncated Nuclear Norm Regularization}

\author{\IEEEauthorblockN{Shengke Xue$^{\letter}$, Wenyuan Qiu, Fan Liu, and Xinyu Jin}
\IEEEauthorblockA{College of Information Science and Electronic Engineering, Zhejiang University, Hangzhou, China\\
Email: \{xueshengke, qiuwenyuan, flyingliufan\}@zju.edu.cn, jinxinyuzju@gmail.com}
}


\maketitle

\begin{abstract}
Currently, low-rank tensor completion has gained cumulative attention in recovering incomplete visual data whose partial elements are missing. By taking a color image or video as a three-dimensional (3D) tensor, previous studies have suggested several definitions of tensor nuclear norm. However, they have limitations and may not properly approximate the real rank of a tensor. Besides, they do not explicitly use the low-rank property in optimization. It is proved that the recently proposed truncated nuclear norm (TNN) can replace the traditional nuclear norm, as a better estimation to the rank of a matrix. Thus, this paper presents a new method called the tensor truncated nuclear norm (T-TNN), which proposes a new definition of tensor nuclear norm and extends the truncated nuclear norm from the matrix case to the tensor case. Beneficial from the low rankness of TNN, our approach improves the efficacy of tensor completion. We exploit the previously proposed tensor singular value decomposition and the alternating direction method of multipliers in optimization. Extensive experiments on real-world videos and images demonstrate that the performance of our approach is superior to those of existing methods. 
\end{abstract}


%
\IEEEpeerreviewmaketitle

\section{Introduction}

Recovering missing values in high dimensional data has become increasingly attractive in computer vision. Exploring the intrinsic low-rank nature of incomplete data with limited observed elements has been widely applied in various applications, e.g., motion capture \cite{Hu2017-Motion}, image alignment \cite{Song2016-Image}, and image classification \cite{Xue2017-Robust}.

Since real visual data (e.g., an image) lies in a low dimensional space, estimating missing values in images used to be considered as a low-rank matrix approximation problem \cite{Candes2009-Exact}. The nuclear norm of a matrix is usually adopted to replace the rank function, which is non-convex and NP-hard in general. Recent studies show that the nuclear norm is suitable for a number of low-rank optimization problems. However, Hu \etal \cite{Hu2013-Accurate} clarified that the nuclear norm may not approximate well to the rank function, since it simultaneously minimizes all of the singular values.

In \cite{Hu2013-Accurate}, the truncated nuclear norm regularization (TNNR) was proposed as a more accurate and much tighter substitute to the rank function, compared with the traditional nuclear norm. It apparently improves the efficacy of image recovery. Specifically, TNNR ignores the largest $r$ singular values of data and attempts to minimize the smallest $\min(m,n)-r$ singular values, where $m \times n$ is the dimension of the two-dimensional data, and $r$ is defined as the number of truncated values. Various studies were derived from it. Liu \etal \cite{Liu2016-Truncated} devised a weighted TNNR using a gradient descent scheme, which further accelerated the speed of its algorithm. Lee and Lam \cite{Lee2016-Computationally} exploited the ghost-free high dynamic range imaging from irradiance maps by adopting the TNNR method.

Moreover, most of existing low-rank matrix approximation approaches deal with the input data in the two-dimensional fashion. Specifically, algorithms are applied on each channel individually and then the results are combined together, when recovering a color image. It reveals an explicit deficiency that the structural information between channels is not considered. Hence, recent studies regard a color image as 3D data and formulate it as a low-rank tensor completion problem.

As an elegant extension of the matrix case, tensor completion earns much interests gradually. However, the definition of the nuclear norm of a tensor is a complicated issue, since it cannot be intuitively derived from the matrix case. Several versions of tensor nuclear norm have been proposed but they are quite different from each other. Liu \etal \cite{Liu2013-Tensor} first proposed a tensor completion approach based on the sum of matricized nuclear norms (SMNN) of the tensor, which is defined as the following formulation:
\begin{equation}
\min_{\tensor{X}} \ \sum_{i=1}^k \alpha_i \norm{\tensor{X}_{[i]}}_* \ \ \text{s.t.} \ \ \tensor{X}_{\bm{\Omega}} = (\tensor{X}_0)_{\bm{\Omega}} , \label{eq:SNN}
\end{equation}
where $\tensor{X}_{[i]}$ denotes the matrix of the unfolded tensor along the $i$th dimension (also known as the mode-$i$ matricization of $\tensor{X}$), $\alpha_i > 0$ is a constant that satisfies $\sum_{i=1}^n \alpha_i = 1$, $\tensor{X}_0$ is the initial incomplete data, and $\bm{\Omega}$ is the set of positions with respect to known elements. But no theoretical analysis interprets that the nuclear norm of each matricization of $\tensor{X}$ is plausible because the spacial structure may lose by the matricization. Moreover, it is unclear  to choose the optimal values of $\alpha_i$ \cite{Zhang2017-Exact}, though they control the weights of $k$ norms in problem \eqref{eq:SNN}. Generally, $\alpha_i$'s are determined empirically.

Recently, a novel tensor decomposition scheme, defined as tensor singular value decomposition (t-SVD), was suggested in \cite{Misha2013-Third-Order}. Based on the new definition of tensor-tensor product, the t-SVD holds some properties that are similar to the matrix case. Therefore, Zhang \etal \cite{Zhang2014-Novel} defined their tubal nuclear norm (Tubal-NN) as the sum of nuclear norms of all frontal slices in the Fourier domain and declared that it is a convex relaxation to the $\ell_1$ norm of the tensor rank. It is formulated as the following problem:
\begin{equation} \label{eq:SFNN}
\min_{\tensor{X}} \ \sum_{i=1}^{n_3} \norm{\fmat{X}^{(i)}}_* \ \ \text{s.t.} \ \ \tensor{X}_{\bm{\Omega}} = (\tensor{X}_0)_{\bm{\Omega}} , 
\end{equation}
where $\fmat{X}^{(i)}$ is defined in Section~\ref{sec:notations}. Semerci \etal \cite{Semerci2014-Tensor-Based} applied this model to multienergy computed tomography images and obtained promising effects of reconstruction. However, it does not explicitly use the low-rank property in optimization and requires numerous iterations to converge. Since the t-SVD is a sophisticated function, the overall computational cost of \eqref{eq:SFNN} will be highly expensive.

To this end, this paper proposes a new approach called the tensor truncated nuclear norm (T-TNN). Based on the t-SVD, we define that our tensor nuclear norm equals to the sum of all singular values in an f-diagonal tensor, which is extended directly from the matrix case. Furthermore, we prove that it can be computed efficiently in the Fourier domain. 
To further take the advantage of TNNR, our T-TNN method generalizes it to the 3D case. In accordance with common strategies, we adopt the universal alternating direction method of multipliers (ADMM) \cite{Lin2011-Linearized} to solve our optimization. Experimental results show that our approach outperforms  previous methods.



\section{Notations and Preliminaries} \label{sec:notations}


In this paper, tensors are denoted by boldface Euler script letters, e.g., $\tensor{A}$. Matrices are denoted by boldface capital letters, e.g., $\bm{A}$. Vectors are denoted by boldface lowercase letters, e.g., $\bm{a}$. Scalars are denoted by lowercase letters, e.g., $a$. $\bm{I}_n$ denotes an $n \times n$ identity matrix. $\mathbb{R}$ and $\mathbb{C}$ are the fields of real number and complex number, respectively. For three-order tensor $\tensor{A} \Cdim{n_1 \times n_2 \times n_3}$, its ($i,j,k$)th element is denoted as $\tensor{A}_{ijk}$. The $i$th horizontal, lateral, and frontal slices of  $\tensor{A}$ are denoted by the Matlab notations $\tensor{A}(i,:,:)$, $\tensor{A}(:,i,:)$, and $\tensor{A}(:,:,i)$, respectively. In addition, the frontal slice $\tensor{A}(:,:,i)$ is concisely denoted as $\bm{A}^{(i)}$. The inner product of matrices $\bm{A}$ and $\bm{B}$ is defined as $\langle \bm{A}, \bm{B} \rangle \triangleq \text{tr} (\bm{A}^\transpose \bm{B})$, where $\bm{A}^\transpose$ is the conjugate transpose of $\bm{A}$ and $\text{tr}(\cdot)$ denotes the trace function. The inner product of $\tensor{A}$ and $\tensor{B}$ in $\mathbb{C}^{n_1 \times n_2 \times n_3}$ is defined as $\langle \tensor{A}, \tensor{B} \rangle \triangleq \sum_{i=1}^{n_3} \langle \bm{A}^{(i)}, \bm{B}^{(i)} \rangle $. The trace of $\tensor{A}$ is defined as $\text{tr}(\tensor{A}) = \sum_{i=1}^{n_3} \trace{\bm{A}^{(i)}}$.

Several norms of a tensor are introduced. The $\ell_1$ norm is defined as $\norm{\tensor{A}}_1 \triangleq \sum_{ijk} |\tensor{A}_{ijk}|$. The infinity norm is defined as $\norm{\tensor{A}}_\infty \triangleq \max_{ijk} |\tensor{A}_{ijk}|$. The Frobenius norm is denoted as $\norm{\tensor{A}}_\fro \triangleq \sqrt{\sum_{ijk}|\tensor{A}_{ijk}|^2}$. The above norms are consistent with the definitions in vectors or matrices. The matrix nuclear norm is defined as $\norm{\bm{A}}_* \triangleq \sum_{i} \sigma_i (\bm{A})$, where $\sigma_i (\bm{A})$ is the $i$th largest singular value of $\bm{A}$.


For $\tensor{A} \Rdim{n_1 \times n_2 \times n_3}$, by using the Matlab notation, we define $\ftensor{A} \triangleq \textsf{fft}(\tensor{A},[\,],3)$, which is the discrete Fourier transformation of $\tensor{A}$ along the third dimension. Likewise, we can compute $\tensor{A} \triangleq \textsf{ifft}(\ftensor{A},[\,],3)$ through the inverse $\textsf{fft}$ function. We define $\fmat{A}$ as a block diagonal matrix, where each frontal slice $\fmat{A}^{(i)}$ of $\ftensor{A}$ lies on the diagonal in order, i.e.,
\begin{equation}
\fmat{A} \triangleq \textsf{bdiag}(\ftensor{A}) \triangleq 
\begin{bmatrix}
\fmat{A}^{(1)} &                    &        &                      \\
& \fmat{A}^{(2)} &        &                      \\
&                    & \ddots &                      \\
&                    &        & \fmat{A}^{(n_3)}
\end{bmatrix} .
\end{equation}

The block circulant matrix of tensor $\tensor{A}$ is defined as
\begin{equation} \label{eq:bcirc}
\textsf{bcirc}(\tensor{A}) \triangleq 
\begin{bmatrix}
\bm{A}^{(1)}   & \bm{A}^{(n_3)}   & \cdots & \bm{A}^{(2)} \\
\bm{A}^{(2)}   & \bm{A}^{(1)}     & \cdots & \bm{A}^{(3)} \\
\vdots         & \vdots           & \ddots & \vdots       \\
\bm{A}^{(n_3)} & \bm{A}^{(n_3-1)} & \cdots & \bm{A}^{(1)}
\end{bmatrix} .
\end{equation}

Here, a pair of folding operators are defined as follows
\begin{equation} \label{eq:unfold}
\textsf{unfold}(\tensor{A}) \triangleq 
\begin{bmatrix}
\bm{A}^{(1)} \\
\bm{A}^{(2)} \\
\vdots \\
\bm{A}^{(n_3)} 
\end{bmatrix} \! , \
\textsf{fold}(\textsf{unfold}(\tensor{A})) \triangleq \tensor{A} \, .
\end{equation}

\begin{definition}[tensor product] \label{def:tensor_product} \cite{Misha2013-Third-Order}
	With $\tensor{A} \Rdim{n_1 \times n_2 \times n_3}$ and $\tensor{B} \Rdim{n_2 \times n_4 \times n_3}$,  the tensor product $\tensor{A} * \tensor{B}$ is defined as a tensor with size $n_1 \times n_4 \times n_3$, i.e.,
	\begin{equation}
	\tensor{A} * \tensor{B} \triangleq \textsf{fold} (\textsf{bcirc} (\tensor{A}) \cdot \textsf{unfold} (\tensor{B}) ) .
	\end{equation}
	The tensor product is similar to the matrix product except that the multiplication between elements is superseded by the circular convolution.
	Note that the tensor product simplifies to the standard matrix product when $n_3 = 1$.
\end{definition}

\begin{definition}[conjugate transpose] \cite{Misha2013-Third-Order}
	If tensor $\tensor{A} \Rdim{n_1 \times n_2 \times n_3}$, the conjugate transpose of which is defined as $\tensor{A}^{\transpose} \Rdim{n_2 \times n_1 \times n_3}$. It is obtained by conjugate transposing each frontal slice and then reversing the order of transposed frontal slices 2 through $n_3$:
	\begin{equation}
	\begin{aligned}
	\big( \tensor{A}^{\transpose} \big)^{(1)} & \triangleq \big( \tensor{A}^{(1)} \big)^\transpose, \\ 
	\big( \tensor{A}^{\transpose} \big)^{(i)} & \triangleq \big( \tensor{A}^{(n_3 + 2 - i)} \big)^\transpose, \ i = 2, \ldots, n_3.
	\end{aligned}
	\end{equation}
\end{definition}

\begin{definition}[identity tensor] \cite{Misha2013-Third-Order}
	Let $\tensor{I} \Rdim{n \times n \times n_3}$ be an identity tensor whose first frontal slice $\bm{I}^{(1)}$ is an $n \times n$ identity matrix and the other slices are zero.
\end{definition}

\begin{definition}[orthogonal tensor] \cite{Misha2013-Third-Order}
	Define tensor $\tensor{Q} \Rdim{n \times n \times n_3}$ is orthogonal if
	\begin{equation}
	\tensor{Q} * \tensor{Q}^{\transpose} \triangleq \tensor{Q}^{\transpose} * \tensor{Q} \triangleq \tensor{I}.
	\end{equation}
\end{definition}

\begin{definition}[f-diagonal tensor] \cite{Misha2013-Third-Order}
	Tensor $\tensor{A}$ is called f-diagonal if each frontal slice $\bm{A}^{(i)}$ is a diagonal matrix.
\end{definition}

\begin{figure}
	\centering
	\includegraphics[scale=0.28]{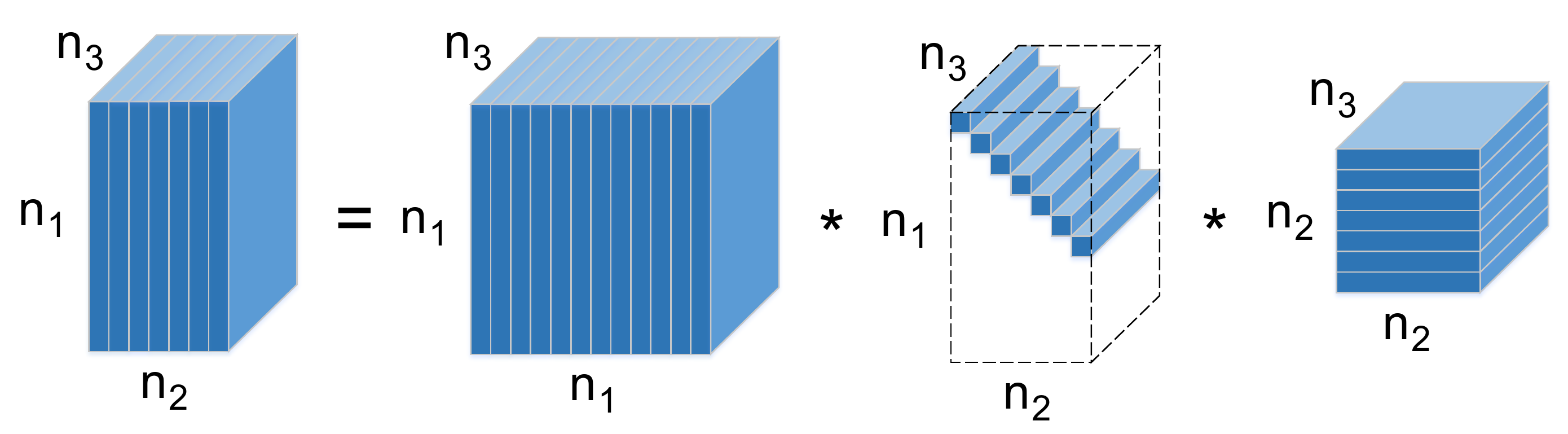}
	\caption{Illustration of the t-SVD of an $n_1 \times n_2 \times n_3$ tensor}
	\label{fig:t-SVD}
\end{figure}

\begin{theorem}[tensor singular value decomposition] \label{the:t-SVD} \cite{Misha2013-Third-Order}
	Tensor $\tensor{A} \Rdim{n_1 \times n_2 \times n_3}$  can be decomposed as
	\begin{equation}
	\tensor{A} \triangleq \tensor{U} * \tensor{S} * \tensor{V}^{\transpose},
	\end{equation}
	where $\tensor{U} \Rdim{n_1 \times n_1 \times n_3}$ and $\tensor{V} \Rdim{n_2 \times n_2 \times n_3}$ are orthogonal, and $\tensor{S} \Rdim{n_1 \times n_2 \times n_3}$ is an f-diagonal tensor.
\end{theorem}
Fig.~\ref{fig:t-SVD} illustrates the t-SVD. However, t-SVD can be efficiently performed on the basis of matrix SVD in the Fourier domain. It comes from an important property that the block circulant matrix can be transformed to a block diagonal matrix in the Fourier domain, i.e.,
\begin{equation}
(\bm{F}_{n_3} \otimes \bm{I}_{n_1}) \cdot \textsf{bcirc} (\tensor{A}) \cdot (\bm{F}_{n_3}^\transpose \otimes \bm{I}_{n_2}) = \fmat{A},
\end{equation}
where $\bm{F}_{n_3}$ denotes the $n_3 \times n_3$ discrete Fourier transform matrix and $\otimes$ denotes the Kronecker product. Note that the matrix SVD can be performed on each frontal slice of $\ftensor{A}$, i.e., $\fmat{A}^{(i)} = \fmat{U}^{(i)} \fmat{S}^{(i)} \fmat{V}^{(i)\transpose}$, where $\fmat{U}^{(i)}$, $\fmat{S}^{(i)}$, and $\fmat{V}^{(i)}$ are the frontal slices of  $\,\ftensor{U}$, $\ftensor{S}$, and $\ftensor{V}$, respectively. More briefly, $\fmat{A} = \fmat{U} \fmat{S} \fmat{V}^\transpose$. Using the {\sf ifft} function along the third dimension, we have $\tensor{U} = \textsf{ifft}(\ftensor{U},[\,],3)$, $\tensor{S} = \textsf{ifft}(\ftensor{S},[\,],3)$, and $\tensor{V} = \textsf{ifft}(\ftensor{V},[\,],3)$.

\begin{definition}[tensor tubal rank and nuclear norm] \label{def:tnn} 
	For tensor $\tensor{A} \Rdim{n_1 \times n_2 \times n_3}$, its t-SVD is $\tensor{U} * \tensor{S} * \tensor{V}^{\transpose}$. The tensor tubal rank of $\tensor{A}$ is defined as the maximum rank among all frontal slices of the f-diagonal $\tensor{S}$, i.e., $\max_{i}\,\text{rank}(\bm{S}^{(i)})$. Our tensor nuclear norm $\norm{\tensor{A}}_*$ is defined as the sum of singular values of all frontal slices of the f-diagonal $\tensor{S}$, i.e.,
	\begin{equation}
		\norm{\tensor{A}}_* \triangleq \trace{\tensor{S}} = \sum_{i=1}^{n_3} \trace{\bm{S}^{(i)}} .
	\end{equation} 
\end{definition}
Note that our tensor nuclear norm becomes standard matrix nuclear norm when $n_3 = 1$. Thus, our tensor nuclear norm can be considered as a direct extension from the matrix case to the tensor case. 

Due to the \textsf{fft} function in the third dimension, we exploit the symmetric property that the trace of tensor product $\tensor{A} * \tensor{B}$ equals to the trace of the product of $\fmat{A}^{(1)}$ and $\fmat{B}^{(1)}$, which are the first frontal slices of  $\ftensor{A}$ and $\ftensor{B}$ in the Fourier domain, i.e.,
\begin{equation}
\trace{ \tensor{A} * \tensor{B} } = \trace{ \fmat{A}^{(1)} \fmat{B}^{(1)} } . \label{eq:fft_property}
\end{equation}
The proof is provided in our arXiv preprint paper. Inspired by \eqref{eq:fft_property}, we further simplify our tensor nuclear norm as follows
\begin{equation}
	\norm{\tensor{A}}_*  \triangleq \trace{\tensor{S}} =  \trace{\fmat{S}^{(1)}} = \norm{\fmat{A}^{(1)}}_* .
\end{equation}
This suggests that our tensor nuclear norm can be efficiently computed by one matrix SVD in the Fourier domain, instead of using the sophisticated t-SVD to obtain $\tensor{S}$.

Our definition is different from those of previous work \cite{Zhang2017-Exact, Lu2016-Tensor}, which are also calculated in the Fourier domain. The tubal nuclear norm in \cite{Zhang2017-Exact} comprises computing each frontal slice of $\ftensor{S}$. Similarly, Lu \etal \cite{Lu2016-Tensor} further proved that the tubal nuclear norm can be computed by the nuclear norm of the block circulant matrix of a tensor with factor $1/n_3$, i.e., $\norm{\tensor{A}}_* = \frac{1}{n_3} \norm{\textsf{bcirc}(\tensor{A})}_*$. However, $\textsf{bcirc}(\tensor{A}) \Rdim{n_1 n_3 \times n_2 n_3}$ requires a huge amount of memory if $n_1$, $n_2$, or $n_3$ is large. This makes the matrix SVD slower.

\begin{definition}[singular value thresholding] \label{def:SVT}
	~Assume the t-SVD of tensor $\tensor{X} \Rdim{n_1 \times n_2 \times n_3}$ is $\tensor{U} * \tensor{S} * \tensor{V}^{\transpose}$. The singular value thresholding (SVT) operator, denoted as $\mathcal{D}_\tau$, is applied on each frontal slice of the f-diagonal tensor $\ftensor{S}$: 
	\begin{equation}
	\begin{gathered}
	\mathcal{D}_\tau (\tensor{X}) \triangleq \tensor{U} * \mathcal{D}_\tau (\tensor{S}) * \tensor{V}^{\transpose}, \ \mathcal{D}_\tau (\tensor{S}) \triangleq \textsf{ifft} ( \mathcal{D}_\tau (\ftensor{S}) ), \\
	\mathcal{D}_\tau (\fmat{S}^{(i)}) \triangleq \text{diag} \big( \max\{\sigma_t - \tau, 0\}_{1 \leq t \leq r} \big), \\
	i = 1, 2, \ldots, n_3.
	\end{gathered}		
	\end{equation}
\end{definition}

\section{Tensor Truncated Nuclear Norm} \label{sec:tnnr}

\subsection{Problem Formulation}

With tensor $\tensor{X} \Rdim{n_1 \times n_2 \times n_3}$, we define our tensor truncated nuclear norm $\norm{\tensor{X}}_r$ as follows
\begin{equation} \label{eq:norm_X_r}
\begin{aligned}
\norm{\tensor{X}}_r & \triangleq \norm{\fmat{X}^{(1)}}_r = \sum_{j=r+1}^{\min(n_1, n_2)} \!\! \sigma_j (\fmat{X}^{(1)}) \\
&=  \sum_{j=1}^{\min(n_1, n_2)} \!\! \sigma_j (\fmat{X}^{(1)}) - \sum_{j=1}^{r} \sigma_j (\fmat{X}^{(1)}) .
\end{aligned}
\end{equation}
By using the Theorem~3.1 in \cite{Hu2013-Accurate}, Theorem~\ref{the:t-SVD}, and Definition~\ref{def:tnn}, we can convert \eqref{eq:norm_X_r} to
\begin{align}
 \norm{\tensor{X}}_r &= \norm{\fmat{X}^{(1)}}_* - \max_{\substack{\fmat{A}^{(1)} \fmat{A}^{(1)\transpose} = \bm{I}, \\ \fmat{B}^{(1)} \fmat{B}^{(1)\transpose} = \bm{I}}} \trace{ \fmat{A}^{(1)} \fmat{X}^{(1)} \fmat{B}^{(1)\transpose} } \nonumber \\
&= \norm{\tensor{X}}_* - \max_{\substack{\tensor{A} * \tensor{A}^\transpose = \tensor{I}, \\ \tensor{B} * \tensor{B}^\transpose = \tensor{I}}} \trace{ \tensor{A} * \tensor{X} * \tensor{B}^\transpose } , \label{eq:tnnr_X_r}
\end{align}
where $\tensor{A}$ and $\tensor{B}$ are generated from the t-SVD of $\tensor{X}$. Define the operator of selecting the first $r$ columns in the second dimension of $\tensor{U}$ and $\tensor{V}$ (using Matlab notation) as follows
\begin{equation}
\tensor{A} \triangleq \tensor{U}(:,1:r,:)^\transpose,\ \tensor{B} \triangleq \tensor{V}(:,1:r,:)^\transpose. \label{eq:generate_A_B}
\end{equation}

Thus, \eqref{eq:tnnr_X_r} can be formulated as the following problem:
\begin{equation} \label{eq:tnnr_with_max}
\begin{aligned}
\min_{\tensor{X}} \ \ & \norm{\tensor{X}}_* - \max_{\substack{\tensor{A}_\ell * \tensor{A}_\ell^\transpose = \tensor{I}, \\ \tensor{B}_\ell * \tensor{B}_\ell^\transpose = \tensor{I}}} \trace{\tensor{A}_\ell * \tensor{X} * \tensor{B}_\ell^\transpose}  \\
\text{s.t.} \, \ \  & \quad \tensor{X}_{\bm{\Omega}} = \tensor{M}_{\bm{\Omega}} ,
\end{aligned}
\end{equation}
where $\tensor{A} \Rdim{n_1 \times r \times n_3}$ and $\tensor{B} \Rdim{n_2 \times r \times n_3}$. It is not easy to directly solve \eqref{eq:tnnr_with_max}, so we divide this optimization into two individual steps. First, we set $\tensor{X}_1 = \tensor{M}_{\bm{\Omega}}$ as the initial value. Assume in the $\ell$th iteration, we update $\tensor{A}_\ell$ and $\tensor{B}_\ell$ as \eqref{eq:generate_A_B} by means of t-SVD (Theorem~\ref{the:t-SVD}). Next, by fixing $\tensor{A}_\ell$ and $\tensor{B}_\ell$, we compute $\tensor{X}_\ell$ through a much simpler problem:
\begin{equation} \label{eq:tnnr_no_max}
\begin{aligned}
\min_{\tensor{X}} \ \ & \norm{\tensor{X}}_* - \trace{\tensor{A}_\ell * \tensor{X} * \tensor{B}_\ell^\transpose}  \\
\text{s.t.} \, \ \  & \quad \tensor{X}_{\bm{\Omega}} = \tensor{M}_{\bm{\Omega}} .  
\end{aligned}
\end{equation}
The detail of solving \eqref{eq:tnnr_no_max} is provided in the next subsection. By alternately taking two steps above, the optimization converges to a local minimum of \eqref{eq:tnnr_with_max}. The framework of our method is summarized in Algorithm~\ref{alg:Tensor-TNNR}.

\begin{algorithm}[t] \small
	\caption{Tensor truncated nuclear norm for low-rank tensor completion}
	\label{alg:Tensor-TNNR}
	\begin{algorithmic}[1]
		\REQUIRE $\tensor{M}$, the original incomplete data; $\bm{\Omega}$, the index set of the known elements; $\bm{\Omega}^{\text{c}}$, the index set of the unknown elements. \\
		\hspace{-6.5mm} \textbf{Initialize:} $\tensor{X}_1=\tensor{M}_{\bm{\Omega}}$, $\varepsilon = 10^{-3}$, $\ell = 1$, $L = 50$. \\
		\REPEAT
		\STATE \textbf{Step 1:} given $\tensor{X}_\ell \Rdim{n_1 \times n_2 \times n_3}$,
		\begin{equation}
		[ \, \tensor{U}_\ell,\tensor{S}_\ell,\tensor{V}_\ell \, ]=\text{t-SVD}(\tensor{X}_\ell), \nonumber  \vspace{-0.5ex}
		\end{equation}
		where the orthogonal tensors are
		\begin{align}
		\tensor{U}_\ell \Rdim{n_1 \times n_1 \times n_3}, \ \tensor{V}_\ell \Rdim{n_2 \times n_2 \times n_3}. \nonumber
		\end{align}
		\vspace{-3ex}
		\STATE Compute $\tensor{A}_\ell$ and $\tensor{B}_\ell$ as follows $\left( r \leq \min\{n_1,n_2\} \right)$: 
		\begin{equation}
		\tensor{A}_\ell = \tensor{U}(:,1:r,:)^\transpose,\ \tensor{B}_\ell = \tensor{V}(:,1:r,:)^\transpose. \nonumber 
		\end{equation} 
		\vspace{-3ex}
		\STATE \textbf{Step 2:} solve the problem: 
		\begin{align}
		\tensor{X}_{\ell+1} = &\, \arg\min_{\tensor{X}} \ \norm{\tensor{X}}_* - \trace{\tensor{A}_\ell * \tensor{X} * \tensor{B}_\ell^\transpose} \nonumber \\
		& \quad \ \ \text{s.t.} \quad \ \ \tensor{X}_{\bm{\Omega}} = \tensor{M}_{\bm{\Omega}} . \nonumber 
		\end{align}
		\UNTIL {$\norm{\tensor{X}_{\ell+1} - \tensor{X}_\ell}_\fro \leq \varepsilon$ or $\ell > L$}
		\ENSURE the recovered tensor.		
	\end{algorithmic}
\end{algorithm}

\subsection{Optimization}

Due to the convergence guarantee in polynomial time, the ADMM is widely adopted in solving constrained optimization problems, such as Step 2 in Algorithm \ref{alg:Tensor-TNNR}. 
First, we introduce an auxiliary variable $\tensor{W}$ to relax the objective. Then \eqref{eq:tnnr_no_max} can be rewritten as 
\begin{equation} \label{eq:step_2_tnnr}
\begin{aligned}
\min_{\tensor{X},\tensor{W}} \ & \norm{\tensor{X}}_* - \trace{\tensor{A}_\ell * \tensor{W} * \tensor{B}_\ell^\transpose} \\
\text{s.t.} \ \ & \tensor{X} = \tensor{W}, \ \tensor{W}_{\bm{\Omega}} = \tensor{M}_{\bm{\Omega}} .
\end{aligned}
\end{equation}
The augmented Lagrangian function of \eqref{eq:step_2_tnnr} becomes
\begin{equation}
\begin{aligned}
 L(\tensor{X}, \tensor{W}, \tensor{Y}) &= \norm{\tensor{X}}_* - \trace{\tensor{A}_\ell * \tensor{W} * \tensor{B}_\ell^\transpose} \\
& \quad + \langle \tensor{Y} , \tensor{X} - \tensor{W} \rangle + \frac{\mu}{2} \norm{\tensor{X} - \tensor{W}}_\fro^2,
\end{aligned}
\end{equation}
where $\tensor{Y}$ is the Lagrange multiplier and $\mu > 0$ is the scalar penalty parameter. Let $\tensor{X}_1 = \tensor{M}_{\bm{\Omega}}$, $\tensor{W}_1 = \tensor{X}_1$, and $\tensor{Y}_1 = \tensor{X}_1$ as the initialization. The optimization of \eqref{eq:step_2_tnnr} consists of the following three steps:

\noindent \textbf{Step 1}: Keep $\tensor{W}_k$ and $\tensor{Y}_k$ invariant and update $\tensor{X}_{k+1}$ from $L (\tensor{X}, \tensor{W}_k, \tensor{Y}_k)$:
\begin{align}
&\tensor{X}_{k+1} = \arg\min_{\tensor{X}} \  \norm{\tensor{X}}_* + \frac{\mu}{2} \norm{\tensor{X} - \tensor{W}_k}_\fro^2 +\langle \tensor{Y}_k , \tensor{X} - \tensor{W}_k \rangle \nonumber \\
& \quad \ \ =  \arg\min_{\tensor{X}} \ \norm{\tensor{X}}_* + \frac{\mu}{2} \Big|\Big| \tensor{X} - \Big( \tensor{W}_k - \frac{1}{\mu} \tensor{Y}_k \Big) \Big|\Big|_\fro^2. \hspace{-1ex} \label{eq:argmin_X}
\end{align}
Based on the SVT operator in Definition \ref{def:SVT}, \eqref{eq:argmin_X} can be solved efficiently by
\begin{equation}
\tensor{X}_{k+1} = \mathcal{D}_{\frac{1}{\mu}} \Big( \tensor{W}_k - \frac{1}{\mu} \tensor{Y}_k \Big) .
\end{equation}

\noindent \textbf{Step 2}: By fixing $\tensor{X}_{k+1}$ and $\tensor{Y}_k$, we can solve $\tensor{W}$ through
\begin{align}
\tensor{W}_{k+1} &= \arg\min_{\tensor{W}} \ L (\tensor{X}_{k+1}, \tensor{W}, \tensor{Y}_k) \nonumber \\
&= \arg\min_{\tensor{W}} \ \frac{\mu}{2} \norm{\tensor{X}_{k+1} - \tensor{W}}_\fro^2 - \trace{\tensor{A}_\ell * \tensor{W} * \tensor{B}_\ell^\transpose} \nonumber \\
& \quad + \langle \tensor{Y}_k , \tensor{X}_{k+1} - \tensor{W} \rangle . \label{eq:argmin_W}
\end{align}
Apparently, \eqref{eq:argmin_W} is quadratic with respect to $\tensor{W}$. Therefore, by setting the derivative of \eqref{eq:argmin_W} to zero, we obtain the closed-form solution as follows
\begin{equation}
\tensor{W}_{k+1} = \tensor{X}_{k+1} + \frac{1}{\mu} \left(\tensor{A}_\ell^\transpose * \tensor{B}_\ell + \tensor{Y}_k \right) \! .
\end{equation}
Furthermore, the values of all observed elements should be constant in each iteration, i.e.,
\begin{equation}
\tensor{W}_{k+1} = (\tensor{W}_{k+1})_{\bm{\Omega}^\text{c}} + \tensor{M}_{\bm{\Omega}}.
\end{equation}

\noindent \textbf{Step 3}: Update $\tensor{Y}_{k+1}$ directly through
\begin{equation}
\tensor{Y}_{k+1} = \tensor{Y}_k + \mu (\tensor{X}_{k+1} - \tensor{W}_{k+1}).
\end{equation}

\section{Experiments} \label{sec:experiment}

In this section, we conduct extensive experiments to demonstrate the efficacy of our proposed algorithm. The compared approaches are:
\begin{enumerate}
	\item Low-rank matrix completion (LRMC) \cite{Candes2009-Exact};
	\item Matrix completion by TNNR \cite{Hu2013-Accurate};	
	\item Tensor completion by SMNN \cite{Liu2013-Tensor};
	\item Tensor completion by Tubal-NN \cite{Zhang2014-Novel};
	\item Tensor completion by T-TNN [ours];
\end{enumerate}
The code of our algorithm is available online at {\url{https://github.com/xueshengke/Tensor-TNNR}}.
All experiments are carried out in Matlab R2015b on Windows 10, with an Intel Core i7 CPU @ 2.60 GHz and 12 GB Memory.
Each parameter of compared methods is readjusted to be optimal and the best results are reported. For fair comparison, each number is averaged over 10 separate runs.

In this paper, each algorithm stops when $\norm{\tensor{X}_{k+1}-\tensor{X}_k}_\fro$ is sufficiently small or the maximum number of iterations has reached. Denote $\norm{\tensor{X}}_{\text{rec}}$ as the final output. Let $\varepsilon = 10^{-3}$ and $L = 50$ for our method. Let $\mu = 5 \times 10^{-4}$ to balance the efficiency and the accuracy of our approach. In practice, the real rank of incomplete data is unknown. Due to the absence of prior knowledge to the number of truncated singular values, $r$ is tested from [1, 30] to choose an optimal value for each case manually.

In general, the peak signal-to-noise ratio (PSNR) is a commonly used metric to evaluate the performances of different approaches. It is defined as follows
\begin{align}
\text{MSE} & \triangleq \norm{(\tensor{X}_\text{rec} - \tensor{M})_{\bm{\Omega}^\text{c}}}_\fro^2 / T, \\
\text{PSNR} & \triangleq 10 \times \log_{10} \left( \frac{255^2}{\text{MSE}} \right) ,
\end{align}
where $T$ is the total number of missing elements in a tensor, and we presume the maximum pixel value in $\tensor{X}$ is 255.

\subsection{Video Recovery}

\begin{figure}[t]
	\centering
	\subfloat[ 65\% element loss]{\includegraphics[width=0.21\textwidth]{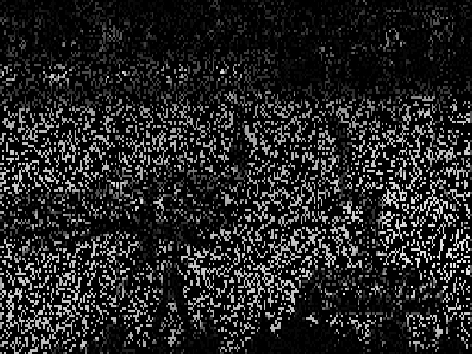}    	\label{subfig:video_incomplete}} \hfil
	\subfloat[PSNR = 17.52, iter = 2084]{\includegraphics[width=0.21\textwidth]{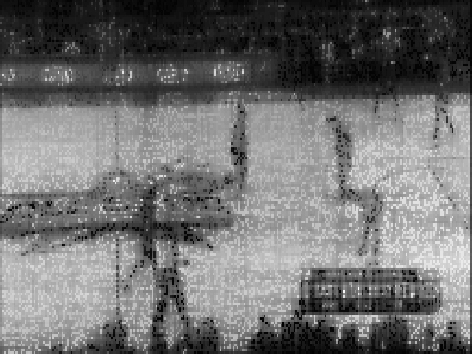}      \label{subfig:video_LRMC}}       \\		\vspace{-1.5ex}	
	\subfloat[PSNR = 18.01, iter = 651 ]{\includegraphics[width=0.21\textwidth]{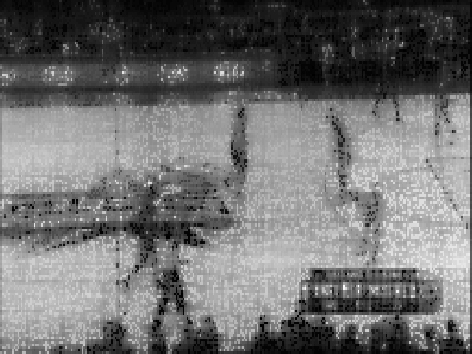}  	\label{subfig:video_TNNR}}       \hfil
	\subfloat[PSNR = 19.49, iter = 161 ]{\includegraphics[width=0.21\textwidth]{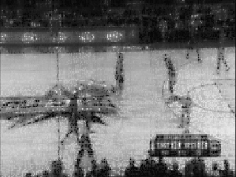}      \label{subfig:video_SMNN}}       \\		\vspace{-1.5ex}
	\subfloat[PSNR = 22.35, iter = 154 ]{\includegraphics[width=0.21\textwidth]{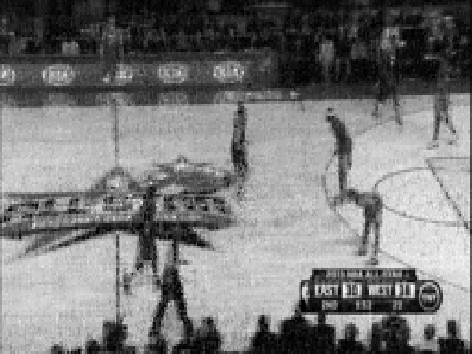}  \label{subfig:video_Tubal-NN}}   \hfil
	\subfloat[PSNR = 24.59, iter = 180 ]{\includegraphics[width=0.21\textwidth]{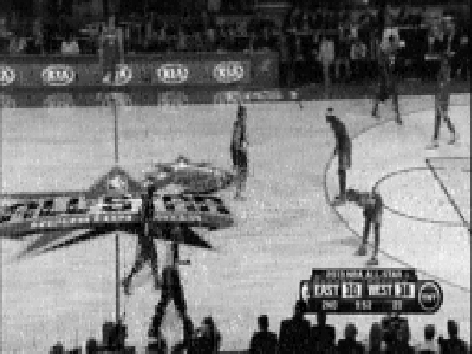}       \label{subfig:video_T-TNN}}       
	\caption{The 20th frame of the basket video reconstructed by five methods: (a) incomplete frame; (b) LRMC; (c) TNNR; (d) SMNN; (e) Tubal-NN; (f) T-TNN } \label{fig:basket_video}
\end{figure}

We consider videos as 3D tensors, where the first and the second dimensions denote space, and the last dimension denotes time. In this experiment, we use a basket video (source: YouTube) with size 144\,$\times$\,256\,$\times$\,40, which is taken from a horizontally moving camera. 65\% elements are randomly lost. Fig.~\ref{subfig:video_incomplete} shows the 20th incomplete frame of the basket video. Our T-TNN method compares to the LRMC, TNNR, SMNN, and Tubal-NN methods. The PSNR, the iteration number, and the 20th frame of the recovered video are reported in Fig.~\ref{fig:basket_video} to validate the effects of five approaches.

Obviously, our T-TNN approach performs best than other methods, though our iteration number is not the least. Fig.~\ref{subfig:video_LRMC} shows that the result of LRMC is the worst, since the matrix completion deals with each frame separately and does not use the spacial structure in the third dimension. Thus, it is not applicable for the tensor case. Beneficial from the truncated nuclear norm, the TNNR (Fig.~\ref{subfig:video_TNNR}) is slightly better than the LRMC. In tensor cases, Fig.~\ref{subfig:video_SMNN} is the worst than the others. This indicates that the SMNN may not be a proper definition for tensor nuclear norm. In addition, Fig.~\ref{subfig:video_T-TNN} is a bit clearer than Fig.~\ref{subfig:video_Tubal-NN}. It implies that our T-TNN method performs best in video recovery.

\subsection{Image Recovery}

In real scenarios, a number of images are  corrupted due to  random loss. 
Because a color image has three channels, we hereby regard it as a 3D tensor rather than separating them.

\begin{figure}[t]
	\centering
	\includegraphics[width=0.47\textwidth]{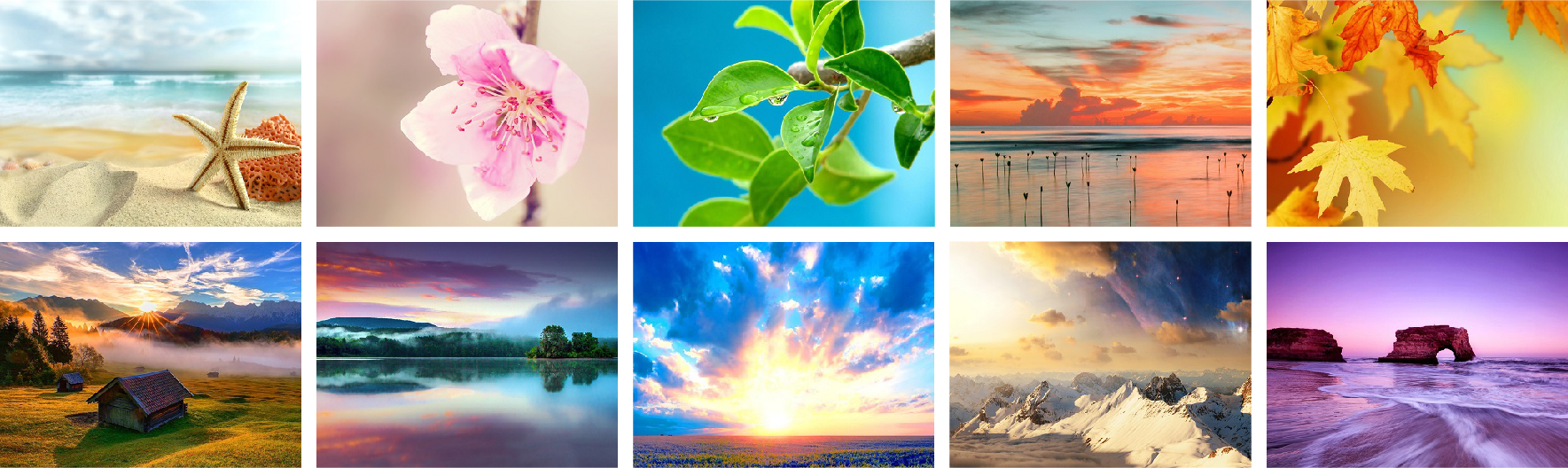}
	\caption{Ten images used in our experiments}
	\label{fig:ten_images}
\end{figure}

\begin{figure}[t]
	\centering
	\subfloat[]{\includegraphics[width=0.21\textwidth]{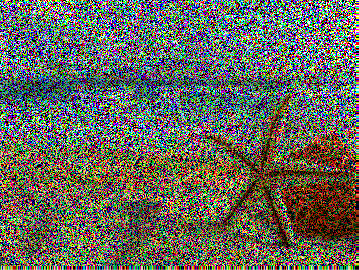} \label{subfig:image_incomplete}} \hfil
	\subfloat[]{\includegraphics[width=0.21\textwidth]{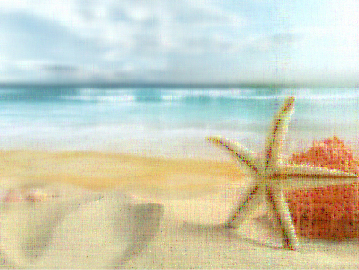}       \label{subfig:image_LRMC}}       \\ \vspace{-1.5ex}	\subfloat[]{\includegraphics[width=0.21\textwidth]{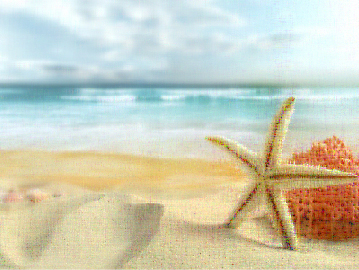}       \label{subfig:image_TNNR}}       \hfil
	\subfloat[]{\includegraphics[width=0.21\textwidth]{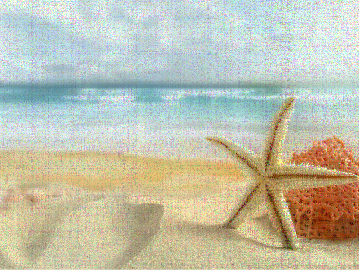}       \label{subfig:image_SMNN}}       \\ \vspace{-1.5ex}
	\subfloat[]{\includegraphics[width=0.21\textwidth]{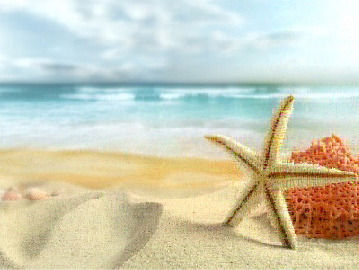}   \label{subfig:image_Tubal-NN}}   \hfil
	\subfloat[]{\includegraphics[width=0.21\textwidth]{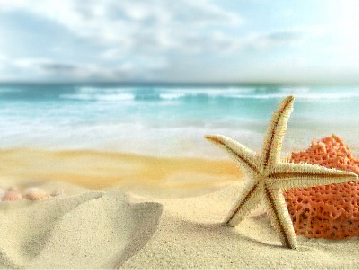}      \label{subfig:image_T-TNN}}      
	\caption{Recovered results of the first image of Fig.~\ref{fig:ten_images} with 50\% random loss by five methods: (a) incomplete image; (b) LRMC; (c) TNNR; (d) SMNN; (e) Tubal-NN; (f) T-TNN }
	\label{fig:recovered_images}
\end{figure}

\begin{figure}[t]
	\centering
	\includegraphics[scale=0.27]{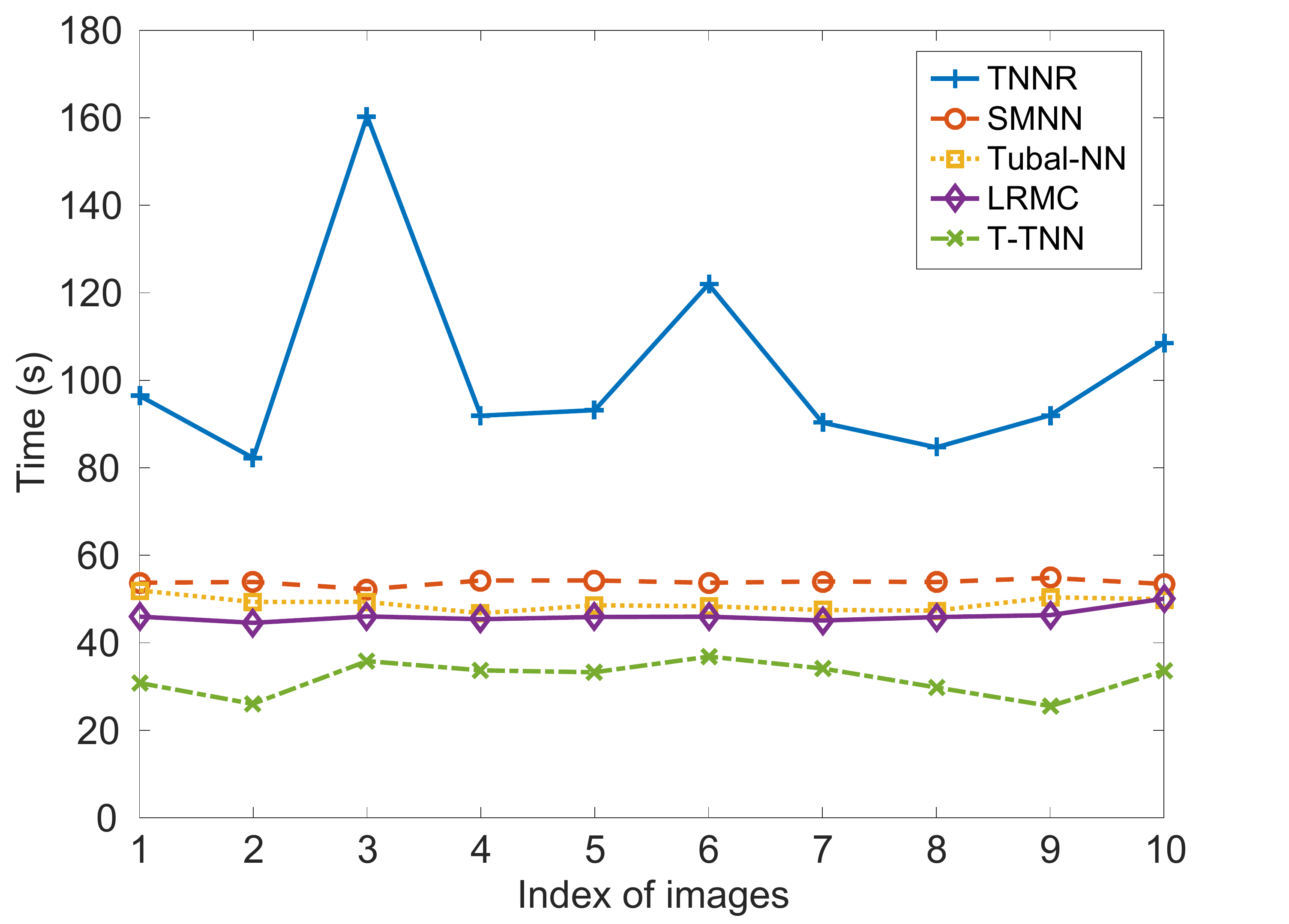}
	\caption{Running time by five methods on ten images (Fig.~\ref{fig:ten_images})}
	\label{fig:time_comparison}
\end{figure}

We use 10 images, as shown in Fig.~\ref{fig:ten_images}, all of which have same size 400\,$\times$\,300. PSNR is adopted to evaluate the effects of image recovery by different algorithms. 50\% pixels in each image are randomly missing. Fig.~\ref{subfig:image_incomplete} presents an example of the incomplete images. 
Under this configuration, our T-TNN approach compares to the LRMC, TNNR, SMNN, and Tubal-NN methods. The PSNR (iteration), the visualized examples of resulting images, and the running time are reported in Table~\ref{tab:PSNR_loss}, Fig.~\ref{fig:recovered_images}, and Fig.~\ref{fig:time_comparison}, respectively.

\begin{table}[t]
	\caption{PSNR of ten recovered images by five methods with 50\% random element loss. Iteration number in parentheses}
	\label{tab:PSNR_loss}
	\centering
		\addtolength{\tabcolsep}{2.5pt}
	\begin{tabular}{|c|c|c|c|c|c|}
		\hline
		       {No.}        &  LRMC  & TNNR  & SMNN  & Tubal-NN &     T-TNN      \\ \hline\hline
		\multirow{2}{*}{1}  & 24.00  & 25.56 & 20.19 &  29.21   & \textbf{29.65} \\
		                    & (1251) & (665) & (343) &  (264)   & \textbf{(165)} \\ \hline
		\multirow{2}{*}{2}  & 26.51  & 29.79 & 22.19 &  32.00   & \textbf{32.55} \\
		                    & (1232) & (541) & (346) &  (258)   & \textbf{(148)} \\ \hline
		\multirow{2}{*}{3}  & 20.95  & 24.33 & 22.23 &  26.37   & \textbf{27.67} \\
		                    & (1274) & (878) & (341) &  (262)   & \textbf{(178)} \\ \hline
		\multirow{2}{*}{4}  & 27.28  & 30.82 & 28.40 &  34.19   & \textbf{35.32} \\
		                    & (1248) & (614) & (351) &  (244)   & \textbf{(181)} \\ \hline
		\multirow{2}{*}{5}  & 25.91  & 28.96 & 24.95 &  30.46   & \textbf{31.20} \\
		                    & (1248) & (656) & (346) &  (246)   & \textbf{(180)} \\ \hline
		\multirow{2}{*}{6}  & 22.21  & 23.23 & 22.95 &  25.54   & \textbf{26.24} \\
		                    & (1251) & (813) & (339) &  (241)   & \textbf{(212)} \\ \hline
		\multirow{2}{*}{7}  & 27.32  & 30.55 & 28.85 &  33.66   & \textbf{34.45} \\
		                    & (1230) & (624) & (345) &  (247)   & \textbf{(188)} \\ \hline
		\multirow{2}{*}{8}  & 23.85  & 26.04 & 22.80 &  29.12   & \textbf{29.93} \\
		                    & (1256) & (573) & (344) &  (249)   & \textbf{(173)} \\ \hline
		\multirow{2}{*}{9}  & 22.68  & 24.17 & 22.53 &  28.22   & \textbf{28.98} \\
		                    & (1261) & (639) & (347) &  (260)   & \textbf{(140)} \\ \hline
		\multirow{2}{*}{10} & 23.52  & 26.60 & 23.40 &  31.59   & \textbf{32.60} \\
		                    & (1262) & (745) & (349) &  (254)   & \textbf{(184)} \\ \hline
	\end{tabular}
\end{table}

Table~\ref{tab:PSNR_loss} presents the PSNR of five approaches applied on ten images (Fig.~\ref{fig:ten_images}) with 50\% random entries lost. Obviously, LRMC requires more than 1000 iterations to converge. Using the truncated nuclear norm, the TNNR notably improves the PSNR of recovery on each image and facilitates the convergence, compared with the LRMC. In tensor cases, the SMNN performs much inferior than the Tubal-NN both in PSNR and iterations, sometimes worse than the LRMC in PSNR. This indicates that the SMNN may not be an appropriate definition for tensor completion. The Tubal-NN holds better PSNR and much faster convergence than the LRMC and SMNN, which implies that  tensor tubal nuclear norm is practical for image recovery. However, our T-TNN is slightly superior in PSNR than the Tubal-NN in all cases and converges much faster.

Fig.~\ref{subfig:image_incomplete} presents the first image of Fig.~\ref{fig:ten_images} with 50\% element loss. In addition, the recovered results by five approaches are illustrated in Figs.~\ref{subfig:image_LRMC}--\ref{subfig:image_T-TNN}, respectively. Obviously, the result of LRMC (Fig.~\ref{subfig:image_LRMC}) contains quite blurry parts and is the worst than the others. With the advantage of the TNN, Fig.~\ref{subfig:image_TNNR} is visually much clearer than Fig.~\ref{subfig:image_LRMC}. However, a certain amount of noise still exists. In tensor cases, Fig.~\ref{subfig:image_Tubal-NN} is pretty clearer than Fig.~\ref{subfig:image_SMNN}, which indicates that the Tubal-NN is much more appropriate than the SMNN. Moreover, the result of SMNN is apparently inferior than the result of TNNR. The result of our method (Fig.~\ref{subfig:image_T-TNN}) is visually competitive to the result of Tubal-NN (Fig.~\ref{subfig:image_Tubal-NN}), both of which contain only a few outliers, compared with the original image.

Fig.~\ref{fig:time_comparison} demonstrates the running time on ten images by five methods. Apparently, the TNNR runs much slower than the others and is erratic on different images (from 82.2\,s to 160.3\,s). Because the ADMM inherently converges slower and consumes more time on SVD operation. The SMNN performs highly stable and spends about 53.0\,s on each image. Similarly, the Tubal-NN and LRMC reveal nearly identical stability as the SMNN, and they are roughly 3.0\,s and 6.0\,s in average faster than the SMNN, respectively. In all cases, our T-TNN method runs quite faster (less than 40\,s) than the Tubal-NN and LRMC, though ours is not sufficiently stable. Because our approach is based on the t-SVD, which achieves  better convergence. Thus, our T-TNN method is efficient for image recovery and is superior to the compared approaches.

\section{Conclusions} \label{sec:conclusion}

This paper proposes the tensor truncated nuclear norm for low-rank tensor completion. Specifically, we present a new definition of tensor nuclear norm, which is a generalization of the standard matrix nuclear norm. In addition, the truncated nuclear norm scheme is integrated in our approach. We adopt previously proposed tensor singular value decomposition and the alternating direction method of multipliers to efficiently solve the problem. Hence, the performance of our method is substantially improved. Experimental results indicate that our approach outperforms the existing methods both in recovering videos and images. Furthermore, the comparisons on running time demonstrate that  our algorithm is further accelerated with the advantage of the truncated nuclear norm. 


\section*{Acknowledgments}


This work was supported by the National Science and Technology Major Project (No.~2013ZX03005013) and the Zhejiang Provincial Natural Science Foundation of China (No.~J20130411).


\bibliographystyle{IEEEtran}
\bibliography{references}

\setcounter{equation}{0}
\renewcommand{\theequation}{\thesection.\arabic{equation}}

\appendix

First, let us recall the definition of the discrete Fourier transformation (DFT) in vectors:
\begin{equation}
	\bar{\bm{x}} \triangleq \bm{F} \bm{x},
\end{equation}
where $\bm{F}$ denotes the Fourier matrix, the $(i,j)$th element of which is defined as $\bm{F}_{ij} = w^{(i-1)(j-1)}$, $w = \exp (-\text{j} 2\uppi / N )$; $\bm{x} = [\bm{x}_1, \bm{x}_2, \ldots, \bm{x}_{N}]^\transpose$ and $\bar{\bm{x}} = [\bar{\bm{x}}_1, \bar{\bm{x}}_2, \ldots, \bar{\bm{x}}_{N}]^\transpose$ are the vectors of time signal and frequency spectrum, respectively. For each element, we have
\begin{equation}
\begin{aligned}
\bar{\bm{x}}_k &= \sum_{n=1}^{N} \bm{x}_n \exp \! \left (-\text{j} \frac{2\uppi (n-1) (k-1)}{N} \right )  \\
&=  \sum_{n=1}^{N} \bm{x}_n w^{(n-1) (k-1)}, \  k = 1,2,\ldots,N.
\end{aligned}
\end{equation}
If $k = 1$, we obtain $\bar{\bm{x}}_1 = \sum_{n=1}^{N} \bm{x}_n$, i.e., the sum of all elements in $\bm{x}$. Then we consider the DFT in the matrix case and the tensor case.

For tensor $\tensor{X} \Rdim{n_1 \times n_2 \times n_3}$, we have $\ftensor{X} \triangleq \textsf{fft}(\tensor{X},[\,],3)$. Note that the fft function runs along the third dimension. Thus, we compute each element in the Fourier domain as follows
\begin{equation}
\begin{gathered}
	\ftensor{X}_{ijk} = \sum_{t=1}^{n_3} \tensor{X}_{ijt} w^{(t-1)(k-1)}, \ w = \exp (-\text{j} 2\uppi / n_3 ), \\
	i = 1,2,\ldots,n_1, \ j = 1,2,\ldots,n_2, \ k = 1,2,\ldots,n_3.
\end{gathered}
\end{equation}
If $k = 1$, we obtain $\ftensor{X}_{ij1} = \sum_{t=1}^{n_3} \tensor{X}_{ijt}$, i.e., the sum of all elements in $\tensor{X}(i,j,:)$. In matrix form, we rewrite it as 
\begin{equation}
	\fmat{X}^{(1)} = \sum_{t=1}^{n_3} \bm{X}^{(t)}. \label{eq:proof_x_fourier}
\end{equation}
By using \eqref{eq:proof_x_fourier}, we can efficiently calculate the trace of a tensor in the Fourier domain as follows
\begin{equation}
	\trace{\tensor{X}} = \sum_{t=1}^{n_3} \trace{\bm{X}^{(t)}} = \text{tr} \bigg( \sum_{t=1}^{n_3} \bm{X}^{(t)} \bigg) = \trace{\fmat{X}^{(1)}}.
\end{equation}

Next, we prove the symmetric property \eqref{eq:fft_property}. Based on \eqref{eq:bcirc}, \eqref{eq:unfold}, and Definition \ref{def:tensor_product}, we have
\begin{align}
&\tensor{A} * \tensor{B} = \textsf{fold} (\textsf{bcirc} (\tensor{A}) \cdot \textsf{unfold} (\tensor{B}) ) \nonumber \\
&= \textsf{fold} \! \left ( \! \begin{bmatrix}
	\bm{A}^{(1)}   & \bm{A}^{(n_3)}   & \cdots & \bm{A}^{(2)} \\
	\bm{A}^{(2)}   & \bm{A}^{(1)}     & \cdots & \bm{A}^{(3)} \\
	\vdots         & \vdots           & \ddots & \vdots       \\
	\bm{A}^{(n_3)} & \bm{A}^{(n_3-1)} & \cdots & \bm{A}^{(1)}
\end{bmatrix}
\! \cdot \!
\begin{bmatrix}
	\bm{B}^{(1)} \\
	\bm{B}^{(2)} \\
	\vdots \\
	\bm{B}^{(n_3)} 
\end{bmatrix} \!
\right ) \nonumber \\
&= \textsf{fold} \! \left ( \begin{bmatrix}
	\sum\limits_{i=1}^{1} \bm{A}^{(2-i)} \bm{B}^{(i)} + \sum\limits_{i=2}^{n_3} \bm{A}^{(n_3+2-i)} \bm{B}^{(i)} \\
	\sum\limits_{i=1}^{2} \bm{A}^{(3-i)} \bm{B}^{(i)} + \sum\limits_{i=3}^{n_3} \bm{A}^{(n_3+3-i)} \bm{B}^{(i)} \\
	\vdots \\
	\sum\limits_{i=1}^{n_3} \bm{A}^{(n_3+1-i)} \bm{B}^{(i)}  \\
\end{bmatrix} \right ) . \label{eq:proof_A*B_fold}
\end{align}
Suppose $\tensor{C} = \tensor{A} * \tensor{B}$ and 
\begin{equation}
	\tensor{C} = \textsf{fold} \! \left (
	\begin{bmatrix}
	\bm{C}^{(1)} \\
	\bm{C}^{(2)} \\
	\vdots \\
	\bm{C}^{(n_3)} 
	\end{bmatrix} \right ) , \label{eq:proof_C_fold}
\end{equation} 
then we obtain the following equality by comparing \eqref{eq:proof_A*B_fold} and \eqref{eq:proof_C_fold}:
\begin{equation}
\begin{cases}
	\bm{C}^{(1)} = \sum\limits_{i=1}^{1} \bm{A}^{(2-i)} \bm{B}^{(i)} + \sum\limits_{i=2}^{n_3} \bm{A}^{(n_3+2-i)} \bm{B}^{(i)}, \\
	\bm{C}^{(2)} = \sum\limits_{i=1}^{2} \bm{A}^{(3-i)} \bm{B}^{(i)} + \sum\limits_{i=3}^{n_3} \bm{A}^{(n_3+3-i)} \bm{B}^{(i)}, \\
	\ \ \vdots \qquad \qquad \qquad \vdots \\
	\bm{C}^{(n_3)} = \sum\limits_{i=1}^{n_3} \bm{A}^{(n_3+1-i)} \bm{B}^{(i)}. \\
\end{cases}
\end{equation}
Using $\trace{\bm{A} \pm \bm{B}} = \trace{\bm{A}} \pm \trace{\bm{B}}$, we compute 
\begin{align}
	& \trace{ \tensor{A} * \tensor{B} } = \text{tr}(\tensor{C}) = \sum_{i=1}^{n_3} \trace{\bm{C}^{(i)}} \nonumber \\
	& = \text{tr} \bigg( \sum\limits_{i=1}^{1} \bm{A}^{(2-i)} \bm{B}^{(i)} + \sum\limits_{i=2}^{n_3} \bm{A}^{(n_3+2-i)} \bm{B}^{(i)} \bigg) \nonumber \\
	& \quad + \text{tr} \bigg( \sum\limits_{i=1}^{2} \bm{A}^{(3-i)} \bm{B}^{(i)} + \sum\limits_{i=3}^{n_3} \bm{A}^{(n_3+3-i)} \bm{B}^{(i)} \bigg) \nonumber \\
	& \quad + \cdots + \text{tr} \bigg( \sum\limits_{i=1}^{n_3} \bm{A}^{(n_3+1-i)} \bm{B}^{(i)} \bigg) \nonumber \\
	& = \text{tr} \bigg( \sum\limits_{i=1}^{1} \bm{A}^{(2-i)} \bm{B}^{(i)} + \sum\limits_{i=2}^{n_3} \bm{A}^{(n_3+2-i)} \bm{B}^{(i)} \nonumber \\
	& \ \ \quad +        \sum\limits_{i=1}^{2} \bm{A}^{(3-i)} \bm{B}^{(i)} + \sum\limits_{i=3}^{n_3} \bm{A}^{(n_3+3-i)} \bm{B}^{(i)} \nonumber \\
	& \ \ \quad + \cdots + \sum\limits_{i=1}^{n_3} \bm{A}^{(n_3+1-i)} \bm{B}^{(i)} \bigg) \nonumber \\
	& = \text{tr} \bigg( \left ( \bm{A}^{(1)} + \bm{A}^{(2)} + \cdots + \bm{A}^{(n_3)} \right ) \bm{B}^{(1)} \nonumber \\
	& \ \ \quad +        \left ( \bm{A}^{(n_3)} + \bm{A}^{(1)} + \cdots + \bm{A}^{(n_3-1)} \right ) \bm{B}^{(2)} + \cdots \nonumber \\
	& \ \ \quad +        \left ( \bm{A}^{(2)} + \bm{A}^{(3)} + \cdots + \bm{A}^{(1)} \right ) \bm{B}^{(n_3)}  \bigg) \nonumber \\
	& = \text{tr} \bigg( \sum\limits_{i=1}^{n_3} \bm{A}^{(i)} \bm{B}^{(1)} + \sum\limits_{i=1}^{n_3} \bm{A}^{(i)} \bm{B}^{(2)} + \cdots + \sum\limits_{i=1}^{n_3} \bm{A}^{(i)} \bm{B}^{(n_3)}  \bigg) \nonumber \\
	& = \text{tr} \bigg( \sum\limits_{i=1}^{n_3} \bm{A}^{(i)} \left ( \bm{B}^{(1)} + \bm{B}^{(2)} + \cdots + \bm{B}^{(n_3)} \right ) \bigg) \nonumber \\
	& = \text{tr} \bigg( \bigg( \sum\limits_{i=1}^{n_3} \bm{A}^{(i)} \bigg) \bigg( \sum\limits_{i=1}^{n_3} \bm{B}^{(i)} \bigg)\bigg)   .
\end{align}
By using \eqref{eq:proof_x_fourier}, we further obtain
\begin{equation}
	\trace{ \tensor{A} * \tensor{B} } = \trace{ \fmat{A}^{(1)} \fmat{B}^{(1)} }.
\end{equation}
Thus, the proof is accomplished. 

\end{document}